\documentclass[runningheads]{llncs}
\usepackage{graphicx}
\usepackage[english]{babel}
\usepackage[utf8]{inputenc}
\usepackage[caption=false,font=footnotesize]{subfig}
\usepackage{caption}
\usepackage{booktabs}
\usepackage{amsmath,amssymb}
\usepackage{hyperref}
\usepackage[table]{xcolor}
\usepackage{collcell}
\usepackage{hhline}
\usepackage{pgf}
\usepackage{multirow}
\usepackage{hyperref}
\usepackage{blindtext}
\usepackage{enumitem}
\usepackage{wrapfig}
\usepackage{soul}
\usepackage[normalem]{ulem}
\usepackage{makecell}
\def\colorModel{hsb} %
\newcommand\ColCell[1]{
  \pgfmathparse{#1<50?1:0}  %
    \ifnum\pgfmathresult=0\relax\color{white}\fi
  \pgfmathsetmacro\compA{0}      %
  \pgfmathsetmacro\compB{#1/100} %
  \pgfmathsetmacro\compC{1}      %
  \edef\x{\noexpand\centering\noexpand\cellcolor[\colorModel]{\compA,\compB,\compC}}\x #1
  } 
\newcolumntype{E}{>{\collectcell\ColCell}m{0.45cm}<{\endcollectcell}}  %

\usepackage[misc,geometry]{ifsym}

\begin{document}
\title{Test Time Transform Prediction for \\ Open Set Histopathological Image Recognition}
\titlerunning{Test Time Transform Prediction for OSHIR}

\author{Adrian Galdran\inst{1,3}$^\textrm{,\Letter}$ \and
Katherine J. Hewitt\inst{2} \and
Narmin L. Ghaffari\inst{2} \and \\
Jakob N. Kather\inst{2} \and
Gustavo Carneiro\inst{3} \and
Miguel A. González Ballester\inst{1,4}}
\authorrunning{A. Galdran et al.}
\institute{BCN Medtech, Dept. of Information and Communication Technologies, Universitat Pompeu Fabra, Barcelona, Spain, \email{\{adrian.galdran,ma.gonzalez\}@upf.edu}
\and
Department of Medicine III, University Hospital RWTH Aachen, Germany, \email{\{khewitt,nghaffarilal,jkather\}@ukaachen.de}
\and
University of Adelaide, Adelaide, Australia, \email{gustavo.carneiro@adelaide.edu}
\and
Catalan Institution for Research and Advanced Studies (ICREA), Barcelona, Spain }

\maketitle              %
\begin{abstract}
Tissue typology annotation in Whole Slide histological images is a complex and tedious, yet necessary task for the development of computational pathology models. 
We propose to address this problem by applying Open Set Recognition techniques to the task of jointly classifying tissue that belongs to a set of annotated classes, \textit{e.g.} clinically relevant tissue categories, while rejecting in test time Open Set samples, \textit{i.e.} images that belong to categories not present in the training set. 
To this end, we introduce a new approach for Open Set histopathological image recognition based on training a model to accurately identify image categories and simultaneously predict which data augmentation transform has been applied. 
In test time, we measure model confidence in predicting this transform, which we expect to be lower for images in the Open Set. 
We carry out comprehensive experiments in the context of colorectal cancer assessment from histological images, which provide evidence on the strengths of our approach to automatically identify samples from unknown categories. Code is released at \url{https://github.com/agaldran/t3po} .
\keywords{Histopathological Image Analysis \and Open Set Recognition}
\end{abstract}

\section{Introduction and Related Work}

Computational pathology has become fertile ground for deep learning techniques, due to several factors like the availability of large scale annotated data coupled with the increase in computational power, or the extremely time-consuming and tedious nature of visual histology examination \cite{van_der_laak_deep_2021,echle_deep_2021,picon_novel_2022}. 
In this context, the advanced pattern recognition capabilities of modern neural networks represents a great match for the challenges posed by digital histopathology.

However, for each new dataset that a practitioner needs to analyze, there is a requirement to annotate large whole slide images, which contain many different tissues, some of them relevant for the task at hand, whereas some others not. 
In this situation, this manual annotation processing can be focused only on the labeling of the relevant tissues. Hence, an algorithm that could automatically disregard data samples outside the set of categories initially labeled by the user would be greatly useful. 
Another plausible scenario arises if the practitioner has labeled all of the tissue typologies that might be of interest to them, but regions of anomalous appearance show up at a later stage. 
These could belong to a category of clinical interest, such as rare disease signs, or simply be acquisition artifacts, but manual review of these findings could be advisable to prevent potential misdiagnosis.
An obvious solution to these problems would be to flag samples in test time for which the computational model is unconfident on its prediction, assuming this could point to atypical data. 
Unfortunately, deep neural networks are known to be incapable of associating anomalous inputs to meaningful low confidence values \cite{guo_calibration_2017}, and there is the need for specific solutions \cite{lee_training_2018}.

A suitable framework to solve the above problems is based on Open Set Recognition (OSR) techniques. 
These are a class of learning algorithms designed to handle the presence in test time of data out of the categories on which a model was trained. 
This is closely related to Out-of-Distribution (OoD) detection; 
for the sake of clarity, we stress that here we follow the definitions given in \cite{vaze_open-set_2021}, and consider OoD detection as the problem of identifying in test time samples that do not belong to the data distribution where the model was trained, without the simultaneous goal of also performing classification on data belonging to known categories. 
For example, a popular approach to OoD detection involves training a model to solve some pretext task for which we know the solution beforehand, \textit{e.g.} predicting the way in which an image has been geometrically transformed \cite{golan_deep_2018}. 
The rationale is that after training, for in-distribution data the model will be able to accurately predict the applied transformation, whereas for OoD data it will most likely fail to recognize it. 
Other common OoD detection methods include  exposing the model to outliers during training \cite{hendrycks_deep_2019}, observing the maximum softmax probability \cite{hendrycks_baseline_2017}, or adding extra branches to the model to account for predictive confidence \cite{devries_learning_2018}. 
These and most other techniques have been proposed in the context of natural images, and it has been shown that they may not translate satisfactorily for OoD detection in medical imaging \cite{berger_confidence-based_2021,zhang_out_2021}.

OSR and OoD detection are also related to Domain Shift/Adaptation (DS/A), the task of training a model to accurately classify data collected in a particular domain, and having the same model generalize to data with the same categories but gathered from a different domain, \textit{e.g.} a second hospital with a different tissue preparation protocol or acquisition device \cite{howard_impact_2021}.
In histopathological image analysis, OoD detection and DS/A have been more studied in recent years than OSR. 
For instance, in \cite{tellez_quantifying_2019} the effect of color augmentation techniques on domain generalization in image classification on slides acquired in 9 different pathology laboratories was analyzed, and in \cite{yamashita_learning_2021} unsupervised style transfer techniques from non-medical data were applied to enhance robustness to domain shift. 
Stacke \textit{et al.} also studied domain shift in histological imaging in \cite{stacke_closer_2019,stacke_measuring_2021}, defining a measure in the space of learned image representations to quantify it and using it to detect data for which a model may struggle to generalize. 
Ensembling techniques are also popular for uncertainty quantification in histological data, and can be put to use for identifying unreliable predictions, which can then be associated to OoD data \cite{thagaard_can_2020}. 
This was proposed for instance in \cite{linmans_efficient_2020}, where multi-head CNNs were shown to be superior to Monte Carlo dropout and deep ensembles for the task of flagging breast histologies containing lymph node tissue showing signs of diffuse large B-cell lymphoma, an anomaly that was not present in the training set.
Self-supervision based on contrastive learning and multi-view consistency has also recently been leveraged for learning robust representations that may enable DA, namely in \cite{ciga_self_2022}. In \cite{bozorgtabar_sood_2021}, the authors used a similar approach to learn representations that could be useful for performing OoD detection under DS.

In this paper we introduce a novel method for OSR on histological images based on recycling information obtained during training regarding the kind of data augmentation operations that are applied online to the training data. 
We conjecture that for images belonging to known categories, a model trained to predict those operations will be more confident in test time, whereas for OoD data the model will be uncertain when solving this pretext task. 
We validate our hypothesis on two popular datasets related to colorectal cancer detection, where our experiments show that the proposed approach can accurately classify images from categories used for training and simultaneously reject clinically uninformative regions in an image without the need to manually label them.

\section{Methodology}
In this section, we introduce basic definitions related to the OSR setting and explain our data augmentation pipeline, which allows us to define transform prediction in a well-posed way. 
We then define our OSR method that jointly classifies in-distribution data and measures confidence in predicting if a data transform operation has been applied in order to declare a sample as OoD.

\subsection{Open Set Recognition - Max over Softmax as a strong baseline}
In an OSR scenario, we start from a labeled training set $\mathcal{C}_\mathrm{train}$ with examples belonging to $N$ known categories $\mathcal{K}=\{k_1,...,k_N\}$, which compose the known, or \textit{Closed Set}. 
However, in test time the classifier encounters samples from an \textit{Open Set} $\mathcal{O}_\mathrm{test}$ with $M$ unknown categories $\mathcal{U}=\{u_1,...,u_M\}$ not seen during training, \textit{i.e.} $\mathcal{D}_\mathrm{test} = \mathcal{C}_\mathrm{test} \cup \mathcal{O}_\mathrm{test}$. 
The goal of an open set classifier is to generate a reliable prediction on $\mathcal{C}_\mathrm{test}$ while also rejecting samples from $\mathcal{O}_\mathrm{test}$.

There exist many approaches to OSR \cite{hendrycks_baseline_2017}. 
However, it has been recently demonstrated in \cite{vaze_open-set_2021} that the simplest of all OSR methods, when optimized so as to maximize closed set accuracy with modern model architectures $U_\theta$ and training techniques, attains state-of-the-art OSR results. 
This baseline method consists of minimizing the cross-entropy loss between one-hot labels $y$ and softmax probabilities $p_\theta(y|x)$ for $x \in \mathcal{C}_\mathrm{train}$, and then define an OSR score as the maximum softmax probability  $S(y \in  \mathcal{C}_\mathrm{test}|x) = \max_{y\in\mathcal{C}} p_\theta(y|x)$, assuming that $U_\theta$ will distribute probabilities with high entropy for unknown classes, resulting in a low $S(y \in  \mathcal{O}_\mathrm{test}|x)$ value.

\subsection{Decoupled Color-Appearance Data Augmentation}
Data augmentation operations (image transforms in computer vision), are a conventional technique to  increase generalization and reduce overfitting when training deep neural networks. 
Recently, learned data augmentation, which learns an optimal transformation policy from a validation set, has gained popularity, with increasingly complex techniques being proposed. 
However, this comes at a noticeable training overhead that has been recently shown to be indeed unnecessary \cite{muller_trivialaugment_2021}: the simpler scheme of randomly selecting, for each optimization step, \textit{a single image transform} (instead of a composition of transforms) from a fixed transform space $\mathcal{T}$, with a variable strength, works remarkably well. 

\begin{figure*}[t]
\centering
\includegraphics[width=0.995\textwidth]{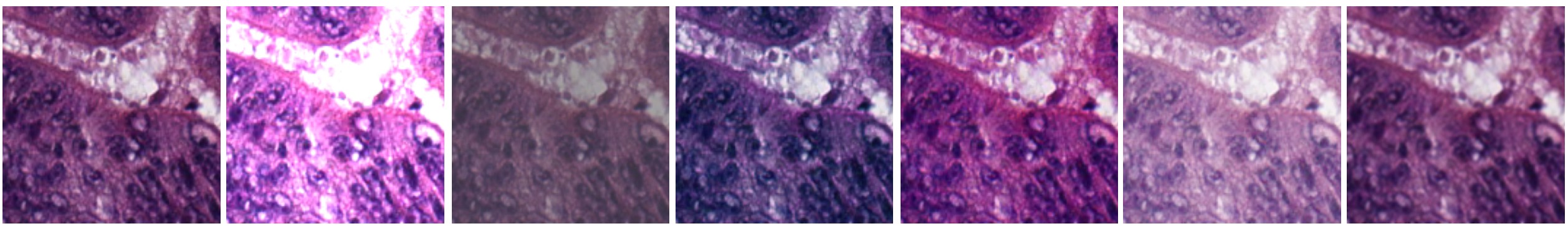}
\caption{Transform space $\mathcal{T}_{\mathrm{app}}$, shown left to right: $\mathcal{T}_{\mathrm{app}}=$\{\texttt{Identity}, \texttt{Brightness}, \texttt{Contrast}, \texttt{Saturation}, \texttt{Hue}, \texttt{Gamma}, \texttt{Sharpness}\}. Our model learns to predict the applied transform during training. In test time, the model only receives un-transformed images, and we measure its confidence on transform prediction.}
\label{fig_transformed_images}
\vspace{-6.35mm}
\end{figure*}

Inspired by \cite{muller_trivialaugment_2021}, we define a data augmentation policy with a single transform at a time, allowing us to pose the auxiliary problem of predicting which transform has been applied to a training sample. 
Also, noting that geometric transforms are hardly predictable on histological data (as opposed to natural images, there is no meaningful notion of top/bottom, rotations, etc.), we decouple geometry from appearance, and define our transform space as $\mathcal{T}=\mathcal{T}_{\mathrm{geom}} \cup \mathcal{T}_{\mathrm{app}}$, where $\mathcal{T}_{\mathrm{geom}}$ contains geometric transformations - rotations, shears, and translations - whereas $\mathcal{T}_{\mathrm{app}}$ contains only color transformations.
These transforms are illustrated and listed in Fig. \ref{fig_transformed_images}; definitions can be found in the standard Python Image Library \url{https://github.com/python-pillow/Pillow}.

\subsection{Test-Time Transform Prediction and Open Set Recognition}
We formulate the training of our model as a joint optimization of two tasks. 
During training, we sample an image $x$ from $\mathcal{C}_\mathrm{train}$, apply a random geometric transform $\tau_g \in \mathcal{T}_{\mathrm{geom}}$, then an appearance transform $\tau_a\in\mathcal{T}_{\mathrm{app}}$, and pass it through a CNN $U_\theta$, which produces an internal representation $x_\theta$. 
This is then sent to the main branch $f_\alpha$, a fully connected layer followed by a softmax operation, which generates a probability of $x$ belonging to a known category from $\mathcal{K}$, but also to an auxiliary branch $g_\beta$ that predicts the actual appearance transform $\tau_a$ that was applied. 
Among these there is the \texttt{Identity} operation, meaning that the model needs to learn what an image $x\in \mathcal{C}$ looks like.

\begin{figure*}[t]
\centering
\includegraphics[width=\textwidth]{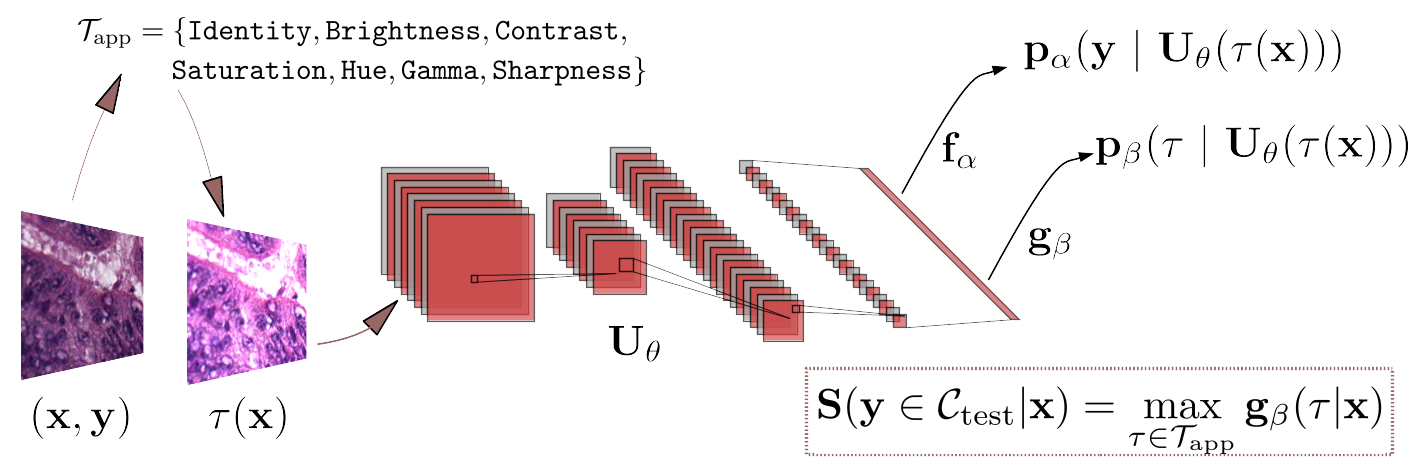}
\caption{Visual scheme of our OSR Test-Time Transform Prediction technique. 
A shared representation $U_\theta(\tau(x))$ is sent to two linear layers, $f_\alpha$ performs Closed Set classification and $g_\beta$ predicts the applied transform $\tau$. In test time, our OSR score is the confidence of the transform prediction branch $g_\beta$ on its prediction.}
\label{fig_scheme}
\vspace{-6.35mm}
\end{figure*}

Finally, in test time, an image $x$ is processed by $U_\theta$ without applying any transform, resulting in a classification score $f_\alpha(U_\theta(x))$, and we define as our OSR score the maximum of softmax probabilities on the transform prediction task:
\begin{equation}
S(y \in  \mathcal{C}_\mathrm{test}|x) = \max_{\tau_a\in\mathcal{T}_{\mathrm{app}}} g_\beta(\tau_a|x).
\end{equation}
In essence, we expect the model to be more confident when predicting the transform on $\mathcal{C}_\mathrm{test}$ than on $\mathcal{O}_\mathrm{test}$. 
Let us note that we could also generate and aggregate predictions on transformed test images (Test-Time Augmentation), although this would induce an inference overhead that we prefer to avoid in this work. 
An illustration of the proposed OSR approach is shown in Fig. \ref{fig_scheme}.

\section{Experimental Analysis}
In this section we introduce our experimental setup: datasets, proposed OSR tasks, and detailed performance evaluation with a discussion on the numerical differences between compared methods, as well as limitations of our technique.

\subsection{Datasets and Open Set Splits}
We evaluate our technique on a clinically meaningful task, namely colorectal cancer (CRC) assessment. 
In this context, tumor tissue composition is heterogeneous, non-stationary, and its study is key to disease prognosis \cite{kather_multi-class_2016}. 
A common technique for CRC monitoring is quantification of tissue configuration by histological evaluation of Hematoxylin and Eosin (H\&E) stained tissue sections.

\begin{figure*}[t]
\centering
\includegraphics[width = 1.0\textwidth]{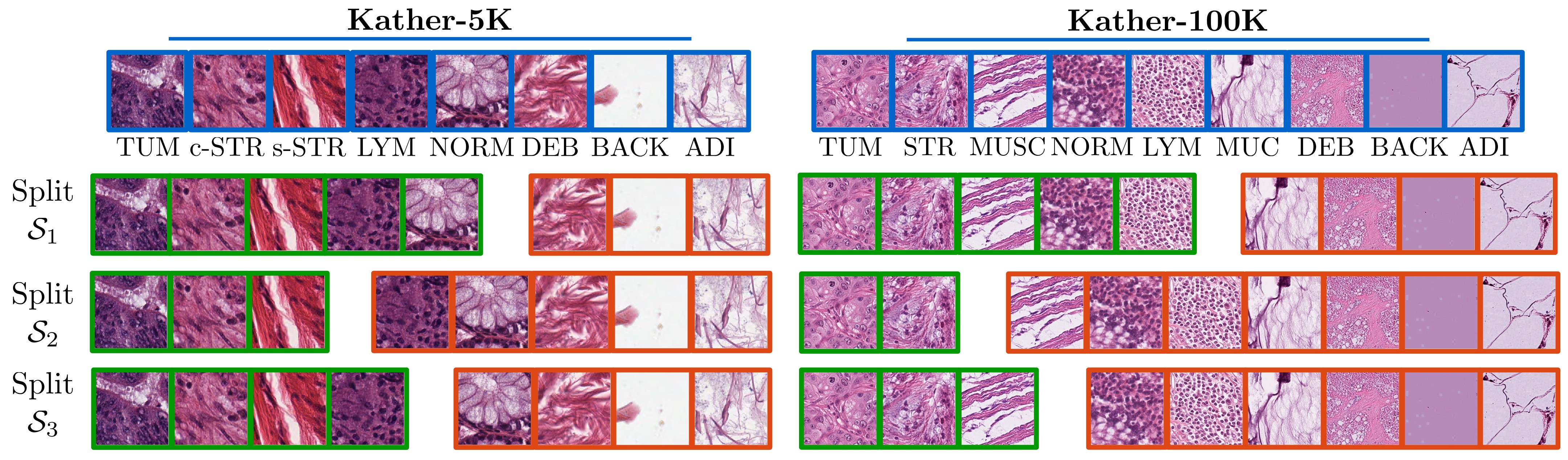}
\caption{Closed (green) and Open (orange) Set partitions defined on the two considered datasets, please see the text for acronym definitions and motivation.}
\label{fig_splits}
\vspace{-6.35mm}
\end{figure*}

We consider two publicly available datasets\footnote{Kather-5k: \url{http://doi.org/10.5281/zenodo.53169} \\Kather-100k: \url{http://doi.org/10.5281/zenodo.1214456}} that enable CRC tissue characterization, referred to as Kather-5k \cite{kather_multi-class_2016} and Kather-100k \cite{kather_predicting_2019}. Examples of images from each tissue type in these datasets are shown in Fig. \ref{fig_splits}. Specifically:
\begin{itemize}[leftmargin=*,noitemsep,topsep=0pt]
\item[$\bullet$] Kather-5k contains 5,000 image patches extracted from 10 tumoral tissue slides with 8 classes: \textit{tumor} epithelium (TUM), \textit{simple stroma} (s-STR), homogeneous tissue, with tumoral and extra-tumoral stroma, but also muscle, \textit{complex stroma} (c-STR), which may contain some immune cells, \textit{immune cell conglomerates} (LYM), \textit{debris} (DEB), which includes necrosis or mucus, \textit{normal} colon mucosa (NORM), \textit{adipose} tissue (ADI), and \textit{background} tissue (BACK). 
Data is balanced, with $150\times 150$ pixel size and $74\mu m$/px resolution.
\item[$\bullet$] Kather-100k is larger, with 100,000 image patches extracted from 86 CRC tissue slides, originally used for overall CRC survival prediction. 
It has 9 different tissue types: \textit{tumour} epithelium (TUM), cancer-related \textit{stroma} (t-STR), smooth \textit{muscle} (MUS), \textit{immune cell conglomerates} (LYM), \textit{debris/necrosis} (DEB), \textit{mucus} (MUC), \textit{normal} colon mucosa (NORM), \textit{adipose} tissue (ADI), and \textit{background} (BACK). 
Note some subtle differences with Kather-5k: the category debris is split into debris/necrosis and mucus; also, stroma is not divided into simple an complex: only tumoral stroma is considered, whereas muscle is a new category.
Data is approximately balanced and color-normalized, with images of size $224\times 224$ and $122\mu$/px resolution. 
\end{itemize}

Next, following expert pathologist's advice, we define several Closed/Open set splits in each dataset, illustrated in Fig. \ref{fig_splits}. 
We first design a split $\mathcal{S}_1$ mimicking the hypothetical situation in which a practitioner decides to label only clinically informative tissue regions, and leaves uninformative regions unlabeled, expecting the OSR model to automatically identify it as part of the Open Set $\mathcal{O}$, while still achieving high accuracy in the Closed Set $\mathcal{C}$.
Note that this is not a trivial task, since necrotic tissue, part of the debris category, can be infiltrated by inflammatory cells, and therefore the lymphocytes class acts as a confounder in the closed set.
To supplement our experimental analysis and understand the weaknesses of OSR systems in this application, we also define two other splits $\mathcal{S}_2$ and $\mathcal{S}_3$.
In $\mathcal{S}_2$ we aim at analyzing if an OSR classifier can classify tumoral regions while rejecting healthy tissue as well as uninformative samples in test time, so we include tumor and stroma patches in the closed set. 
Note that for both datasets there are now some confounders in the Open Set. 
Namely, in the Kather-5k dataset complex stroma images may include some immune cells, but the immune-cell conglomerate category is in the Open Set of $\mathcal{S}_2$. 
On the other hand, in the Kather-100k dataset stroma images do not include immune cells, but the stroma and the muscle categories share a fibrous aspect, and muscle images belong to the Open Set. 
For comparison purposes, in the last split $\mathcal{S}_3$ we move the lymphocites class to the Closed Set, which should result in an easier OSR task at the expense of a more challenging Closed Set classification task, since now $\mathcal{C}$ contains two similar classes.

\subsection{Implementation Details and Performance Evaluation}
We compare Test-Time Transform Prediction (T3PO) with the state-of-the-art ARPL technique \cite{chen_adversarial_2021}, and CE+, the strong baseline proposed recently in \cite{vaze_open-set_2021}, which consists of maximizing the Closed Set accuracy of the classifier and, instead of taking the maximum over the softmax probabilities as the OSR score, use the maximum over the logits, \textit{i.e.} pre-softmax activations of the network. 
We also adopt the MC-Dropout baseline (applying dropout multiple times ($n=32$) in test time and collecting the entropy of the resulting set of softmax probabilities as the OSR score), popular in medical image analysis problems \cite{linmans_efficient_2020}.

Since previous work has shown that relatively small architectures are capable of achieving high accuracy on the two considered datasets, for the sake of quick experimentation we always train a MobileNet V2 network as our backbone \cite{sandler_mobilenetv2_2018}, starting from ImageNet weights. 
Following \cite{kather_multi-class_2016}, we split the data into $70\%$ for training, $15\%$ for validation and early-stopping, and $15\%$ for testing.
In all cases we train with a cyclical learning rate starting at $l=0.01$ and a batch-size of $128$, for $200$ epochs in the Kather-5k dataset. 
Due to the larger amount of training samples, we only train for $20$ epochs in the Kather-100k dataset, which is enough for all models to converge. 
We use the Adam optimizer, monitor the Closed Set accuracy during training, and keep the highest-performing checkpoint. 
After training, we collect model accuracy on the Closed test set, and OSR scores in the Closed and Open test sets.
We perform ten training runs per split and report mean Closed Set accuracy and Closed/Open AUC.

\begin{table}[t]
\renewcommand{\arraystretch}{1.0}	
\centering
\setlength\tabcolsep{10pt}
\caption{Performance averaged over 10 training runs of our approach and other OSR techniques on several Open/Closed splits of the Kather-5k dataset. Best performance is underlined, results within its confidence interval are bold.}
\label{results_kather5k}
\begin{tabular}{l cc cc cc}
    & \multicolumn{2}{c}{\textbf{Split 0}} & \multicolumn{2}{c}{\textbf{Split 1}} & \multicolumn{2}{c}{\textbf{Split 2}} \\
    \cmidrule(lr){2-3} \cmidrule(lr){4-5} \cmidrule(lr){6-7}
                                &  ACC                       &  AUC                       &  ACC                      &  AUC                       & ACC                        &  AUC   \\
\midrule
CE+                             & \textbf{93.03}             & 91.66                      &\underline{\textbf{94.27}} &  82.51                     & 92.88 & 90.02  \\
\midrule
ARPL                            & \textbf{92.84}             & 88.96                      & 92.51                     &  80.28        	           &  \underline{\textbf{93.39}} & 82.39  \\
\midrule
MC-Dropout                      & \underline{\textbf{93.16}} & 91.52                      & \textbf{94.02}            &  82.19         	           &  92.80            & 85.45   \\
\bottomrule
\textbf{T3PO (Ours)}            & \textbf{92.54}             & \underline{\textbf{93.55}} &\underline{\textbf{94.27}} & \underline{\textbf{84.73}} & 91.80             & \underline{\textbf{91.24}} \\
\bottomrule
\end{tabular}
\end{table}

\subsection{Results and Discussion}
Tables \ref{results_kather5k} and \ref{results_kather100k} show the performance of the considered OSR techniques on the Kather-5k and Kather-100k datasets respectively. 
The first split, which in both cases sets out the task of classifying clinically relevant tissue categories, is successfully solved to a high accuracy by all approaches, with no statistically significant difference between our proposed T3PO and the top performer MC-Dropout. 
If we analyze the ability of each method to reject uninteresting data in test time, however, we see that T3PO outperforms the other techniques, by a relatively wide margin in the Kather-5k dataset, in terms of Closed/Open Set AUC, indicating that our method can better identify Open Set data in this case. 

The second and third split in the Kather-5k dataset illustrate a limitation of OSR approaches. 
In the second split, the Open Set contains images from the immune cell category, and immune cells are also present on some images from the complex stroma class, which belongs to the Closed Set. 
This results in a generally lower AUC for all methods, although T3PO continues to outperform other techniques. 
In addition, when we move the immune-cell category from the Open to the Closed Set, we see a noticeable increase in AUC for all methods (and a decrease in accuracy, since two visually similar categories are now in the Closed Set), with T3PO still significantly attaining top performance in Open Set recognition. 
It should be noted that this is achieved at the cost of a modest, but statistically significant decrease in Closed Set accuracy for the third split.

Lastly, the second and third split in the Kather-100k dataset also show a similar phenomenon. 
In this case the muscle class belonging to the Open Set in the second split drives the Closed/Open AUC down for all methods, since it is confounded with the stroma category from the Closed set, and we see that T3PO is among the worst techniques now. 
However, when we move the muscle class into the Closed Set, T3PO increases the AUC by more than 9 points, outperforming all other methods, and losing very little accuracy.

\begin{table}[t]
\renewcommand{\arraystretch}{1.0}	
\centering
\setlength\tabcolsep{10pt}
\caption{Performance averaged over 10 training runs of our approach and other OSR techniques on several Open/Closed splits of the Kather-100k dataset. Best performance is underlined, results within its confidence interval are bold.}
\label{results_kather100k}
\begin{tabular}{l cc cc cc}
    & \multicolumn{2}{c}{\textbf{Split 0}} & \multicolumn{2}{c}{\textbf{Split 1}} & \multicolumn{2}{c}{\textbf{Split 2}} \\
    \cmidrule(lr){2-3} \cmidrule(lr){4-5} \cmidrule(lr){6-7}
                                &  ACC                       &  AUC                		 &  ACC                       &  AUC                        & ACC                        &  AUC   \\
\midrule
CE+                             &  \textbf{99.54}            & 96.50                	 & \underline{\textbf{99.69}} & \textbf{84.59}              & \underline{\textbf{99.62}} &  82.96  \\
\midrule
ARPL                            & 98.88                      & 91.76              	     & \textbf{99.33}             &  78.00        	            &  98.98                     & 79.96   \\
\midrule
MC-Dropout                      & \underline{\textbf{99.57}} & 96.23                	 & \textbf{99.64}             &  \underline{\textbf{84.93}} &  \textbf{99.58}            & 84.52 \\
\bottomrule
\textbf{T3PO (Ours)}            & \textbf{99.46}             & \underline{\textbf{96.57}}&  \textbf{99.66}            &  83.32                      &  \textbf{99.56}            &\underline{\textbf{92.42}} \\
\bottomrule
\end{tabular}
\end{table}

\subsection{Conclusion and Future Work}
We have illustrated how a clinically meaningful task, disregard irrelevant image regions from histological slides without explicitly training a model to discriminate them, can be addressed with OSR techniques. 
We have also introduced T3PO, a new OSR method that outperforms several recent approaches in most cases. 
We have also discussed its limitations, namely T3PO consists of the identification of global image transformations in test time, thereby relying on low-level image characteristics like color and aspect, but not taking full advantage of other semantic cues, which may result in sub-optimal performance. 
We leave the integration of the knowledge of image content into our approach for future work.

\section*{Acknowledgments}
This work was partially supported by a Marie Skłodowska-Curie Global Fellowship (No 892297) and by Australian Research Council grants (DP180103232 and FT190100525).

\bibliographystyle{splncs04}
\bibliography{miccai_osr.bib}

\end{document}